\def\ARXIVVERSION{1}
\documentclass[11pt]{article}

\PassOptionsToPackage{table}{xcolor}
\ifdefined\ARXIVVERSION
  \usepackage[preprint]{acl}
\else
  \usepackage[final]{acl}
\fi

\usepackage{times}
\usepackage{latexsym}
\usepackage[T1]{fontenc}
\usepackage[utf8]{inputenc}
\usepackage{microtype}
\usepackage{inconsolata}

\usepackage{graphicx}
\graphicspath{{figure/}{../figure/}}
\usepackage{amsmath}
\usepackage{amssymb}
\usepackage{adjustbox}

\usepackage{booktabs}
\usepackage{colortbl}
\usepackage{multirow}
\usepackage{subcaption} 
\usepackage{algorithm}
\usepackage{algpseudocode}
\usepackage{float}
\usepackage{enumitem}

\definecolor{PlotBlue}{RGB}{230,240,255}
\definecolor{PlotGreen}{RGB}{235,245,235}
\definecolor{PlotPurple}{RGB}{240,235,255}
\definecolor{HeaderGray}{gray}{0.9}
\definecolor{lightblue}{RGB}{221,235,247}   
\definecolor{graybg}{RGB}{242,242,242}      
\definecolor{gaincolor}{RGB}{34,139,34}     
\definecolor{losscolor}{RGB}{200,0,0}       
\definecolor{deeppink}{RGB}{255, 20, 147}
\definecolor{graygain}{RGB}{128, 128, 128}

\newcommand{\gbf}[1]{\textcolor{gaincolor}{\textbf{#1}}} 
\newcommand{\rtext}[1]{\textcolor{losscolor}{#1}}

\title{Stable-MM-R1: Anchoring Multimodal Reasoning Dynamics via Entropy-Guided Stratification}

\author{
  \textbf{Yimeng Ye}\textsuperscript{1,*},
  \textbf{Shawn Chen}\textsuperscript{1,*},
  \textbf{Wenxuan Huang}\textsuperscript{2},
  \textbf{Manyuan Zhang}\textsuperscript{2}
  \\
  \textbf{Kaituo Feng}\textsuperscript{2},
  \textbf{Zhangquan Chen}\textsuperscript{3},
  \textbf{Jiayu Chen}\textsuperscript{2},
  \textbf{Yucheng Zhou}\textsuperscript{2}
  \\
  \textbf{Yicheng Xiao}\textsuperscript{3},
  \textbf{ZhiYuan Feng}\textsuperscript{3},
  \textbf{Tianyu Shi}\textsuperscript{4}
  \\[1ex]
  \normalfont\normalsize
  \textsuperscript{1}Columbia University
  \quad
  \textsuperscript{2}CUHK
  \quad
  \textsuperscript{3}THU
  \quad
  \textsuperscript{4}McGill University
  \\
  \small\textsuperscript{*}Equal contribution
}

\begin{document}
\maketitle
\begin{abstract} While Reinforcement Learning (RL) effectively incentivizes reasoning in Large Language Models, current pipelines are hindered by training instability and rapid entropy collapse. These limitations often stem from "Rollout Silencing" and low-quality gradient signals in standard sampling procedures. In this work, we propose a robust, data-centric framework to stabilize RL training. We first introduce Potential-Aware Query Mining (PAQM), which filters data dynamically to focus on the "Distillation Zone"---samples with high potential for capability elicitation. Furthermore, we present Hybrid Stratified Replay (HSR), a novel mechanism that restructures batches by stratifying rollouts based on Path Entropy, a rollout-level confidence proxy, and outcome reward. Within each optimization step, HSR reuses current-policy "Stability Anchors" and "Hard Negatives" to construct high-contrast optimization groups, then clears its buffers before the next step. This approach mitigates entropy collapse while improving the utilization of learning signals under limited compute. Our method outperforms strong baselines on complex reasoning tasks, offering a principled solution for stable and efficient RL fine-tuning. \end{abstract}

\section{Introduction}

Multimodal large language models (MLLMs) have recently demonstrated striking progress on complex reasoning tasks, especially when equipped with explicit reasoning traces such as chain-of-thought (CoT) \citep{wei2022cot,chen2026unify}. 
However, reliably \emph{eliciting} and \emph{stabilizing} reasoning behaviors during post-training remains challenging, particularly when we move beyond supervised imitation toward reinforcement learning (RL) with verifiable rewards \citep{ouyang2022instructgpt, chen2026ares, deepseekr1_2025,zhang2025critique,feng2026onethinker}. 
In reasoning-centric settings (e.g., math and diagram reasoning), RL is attractive because correctness can be automatically verified, enabling scalable learning signals without dense human labels \citep{cobbe2021gsm8k, lightman2023prm}. 
Yet practitioners consistently observe that RL training can be brittle: reward curves oscillate, KL/entropy diagnostics exhibit abrupt shifts, and the policy may prematurely collapse into narrow modes that look confident but generalize poorly \citep{yuan2025ppocollapse, dapoyuscale_2025}.

Following recent reasoning LLM practice, we define stable training as a process where (i) reward and benchmark performance improve steadily across steps, and (ii) the model's internal state evolves smoothly without abrupt regime changes, as reflected by diagnostics such as training--inference KL divergence and sequence-level/token-level entropy \citep{yuan2025ppocollapse}. 
Unfortunately, stability is often violated in modern RL pipelines for reasoning.
A prominent failure mode is premature entropy collapse: once the policy discovers a narrow high-reward strategy, the action distribution sharpens too quickly, cutting off exploration and trapping learning in local optima.
This issue becomes more severe in long-CoT reasoning where gradient signal is sparse and delayed, and even PPO-style actor--critic RL can collapse without careful value calibration \citep{schulman2017ppo, yuan2025ppocollapse}. 
Moreover, group-based methods that avoid a learned critic---e.g., GRPO---reduce system complexity and have enabled strong reasoning results \citep{shao2024deepseekmath,chen2026opensearch, deepseekr1_2025, feng2026gen,wu2026reinforcing, huang2026vision, zeng2026vision, fang2026videodeepresearchnextgenerationmultimodaldeepresearch}, but still suffer from batch degeneracy: stochastic group composition can yield near-zero advantage variance (vanishing updates) or noisy baselines (destabilizing updates), especially under limited rollout budgets.

Common stabilization mechanisms are \emph{objective-centric}, including clipped policy updates and KL or entropy control \citep{schulman2017ppo, dapoyuscale_2025}.
While effective in some regimes, these ``soft constraints'' can be sensitive to scaling, reward sparsity, and long-horizon CoT dynamics.
Inspired by recent data-centric RL observations (e.g., advantage collapsing / rollout silencing in multimodal RL fine-tuning) \citep{shuffleR1_2025}, we ask a different question:
\emph{instead of modifying how RL optimizes, can we stabilize training by controlling what data the optimizer sees at each step?}
We argue that a major source of instability is not the RL objective itself, but the stochastic quality of rollouts and optimization groups: easy queries waste compute with near-zero gradients, impossible queries inject high-variance noise, and mixed-quality rollouts within a group can produce unreliable advantage estimates.


To address these challenges, we propose \textbf{Stable-MM-R1}, a data-centric stabilization framework for reasoning RL that structures learning around two complementary axes: outcome (verification reward) and confidence (Path Entropy, defined as a sequence-level surprisal proxy).
Unlike standard methods that rely on stochastic sampling, our approach actively manages the training distribution.
First, we introduce \textit{Potential-Aware Query Mining (PAQM)} to dynamically curate a curriculum, ensuring the model continuously operates within a "Distillation Zone" of learnable queries rather than wasting compute on the trivial or the impossible.
Second, and most crucially, we employ \textit{Hybrid Stratified Replay (HSR)} to engineer the composition of optimization groups. By pairing high-surprisal exploration targets with "Stability Anchors" (mastered samples) and "Hard Negatives" (confident errors), we construct stable, high-contrast optimization groups whenever the corresponding samples are available.
Together, these components are designed to reduce the incidence of zero-signal groups and mitigate premature entropy collapse without modifying the underlying RL objective.

Our main contributions are:
(1) We identify stochastic group degeneracy as one source of instability in reasoning RL, showing from a data-centric perspective how uncontrolled rollout quality can lead to vanishing gradients and premature entropy collapse.
(2) We introduce Potential-Aware Query Mining (PAQM), a dynamic curriculum framework that filters data based on \textit{Pass@K} and \textit{Path Entropy} to concentrate updates within the "Distillation Zone."
(3) We propose Hybrid Stratified Replay (HSR), a novel sampling strategy that reduces the incidence of degenerate groups through the structured inclusion of Stability Anchors and Hard Negatives when the corresponding samples are available, thereby anchoring the policy update dynamics.
(4) We demonstrate that Stable-MM-R1 outperforms the evaluated open-source 7B-scale baselines on average across diverse benchmarks, while maintaining healthy Path-Entropy trajectories that avoid the collapse patterns observed in GRPO and DAPO.

\begin{figure*}[t]
\centering
\includegraphics[width=\textwidth]{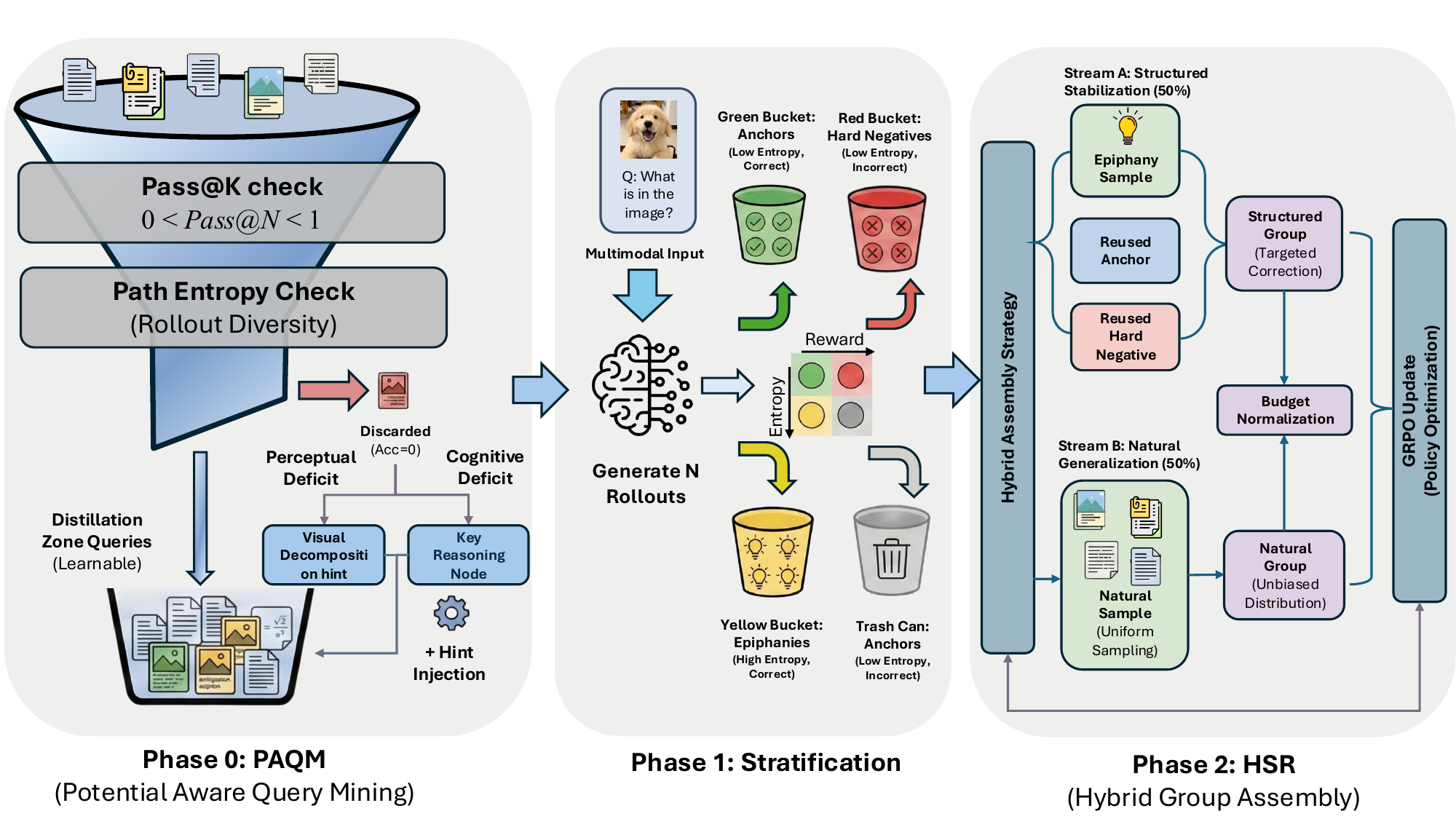}
\caption{Overview of the Stable-MM-R1 framework. 
\textbf{Phase 0 (PAQM):} A dynamic curriculum is constructed via a multi-dimensional filter (Pass@K and Path Entropy). Queries in the "Impossible" zone undergo a \textit{Modality-Decoupled Diagnosis}, receiving targeted \textit{Visual Decomposition} or \textit{Cognitive Key Nodes} to reclaim them into the learnable frontier.
\textbf{Phase 1 (Stratification):} Generated rollouts are diagnosed using Path Entropy and reward, then classified into semantic buffers: \textit{Anchors} (stability), \textit{Hard Negatives} (correction), and \textit{Epiphanies} (exploration), while stochastic \textit{Noise} is discarded.
\textbf{Phase 2 (HSR Group Assembly):} Optimization groups are assembled via a hybrid strategy. \textit{Stream A} (Structured) explicitly reuses Anchors and Hard Negatives to stabilize the gradient variance for Epiphanies, while \textit{Stream B} (Natural) preserves natural on-policy samples. Together, Phases 1 and 2 constitute the complete HSR pipeline.}
\label{fig:main_architecture}
\end{figure*}

\section{Related Works}

\subsection{Multimodal Large Reasoning Models (MLRMs)}
The paradigm of post-training has shifted from supervised imitation to reinforcement learning (RL), driven by the premise that reasoning capabilities emerge from exploration rather than memorization \citep{feng2026video,chen2025advancing}. Pioneering works like DeepSeek-R1 \citep{deepseekr1_2025} and Kimi k1.5 \citep{kimi_k15_2025} have demonstrated that RL with verifiable rewards can incentivize long-chain reasoning (CoT). This trend is rapidly extending to the multimodal domain with models like Qwen2.5-VL \citep{bai2025qwen25vl} and LLaVA-OneVision \citep{llavaonevision}.

Recent specialized frameworks have further expanded this frontier. SATORI-R1 \citep{shen2025satori} integrates spatial grounding with verifiable rewards to incentivize fine-grained visual reasoning by decomposing VQA into verifiable stages. Similarly, Retrv-R1 \citep{zhu2025retrv} introduces a reasoning-driven framework tailored for universal multimodal retrieval. Our work complements these by focusing on the stability of the underlying optimization process itself.

\subsection{Entropy Collapse and Stabilization in RL}
A critical failure mode in reasoning-oriented RL is Entropy Collapse, where the policy prematurely converges to a deterministic distribution \citep{yuan2025ppocollapse}. PPO-style training constrains policy updates through clipping and can additionally use KL or entropy control, but group-based methods such as GRPO still struggle to maintain diversity \citep{schulman2017ppo}.

Recent works have proposed various \textit{objective-centric} modifications to stabilize GRPO. Quantile Advantage Estimation (QAE) \citep{wu2025quantile} replaces the mean baseline with quantile estimates to prevent reward outliers from crushing entropy. Sharpness-Guided GRPO (GRPO-SG) \citep{le2025token} uses probability-based token weights to downweight tokens likely to induce overly large gradients. Similarly, M-GRPO \citep{bai2025mgrpo} utilizes a momentum-updated reference model to anchor the policy target, reducing oscillation. Taking a control-theoretic perspective, EntroPIC \citep{yang2025entropic} applies Proportional-Integral (PI) control to dynamically adjust loss coefficients, explicitly targeting a desired entropy trajectory.

\subsection{Data-Centric Policy Optimization}
An emerging alternative focuses on the \textit{input data distribution} rather than the objective function. Shuffle-R1 \citep{shuffleR1_2025} identifies "rollout silencing" and proposes contrastive batch organization. Concurrently, \citet{tang2025rethinking} emphasize sample polarity, while MMR1 \citep{leng2025mmr1} explores variance-aware sampling.

Beyond selection, recent works explore restructuring the data itself. RoRecomp \citep{li2025rorecomp} enhances reasoning efficiency via rollout response recomposition. Addressing gradient variance directly, \citet{zeng2025shrinking} introduce Shrinkage Baselines for verifiable rewards to improve estimator accuracy. Unlike these approaches, our Stable-MM-R1 introduces Path Entropy, a rollout-level confidence proxy, as a dimension for data curation. By explicitly stratifying samples into stability anchors and exploration targets, we enforce training stability through structured data composition.

\section{Methodology}
\label{sec:method}

In this section, we present Stable-MM-R1, a unified framework designed to stabilize the post-training of multimodal reasoning models. Our approach operates in two stages: first, we curate a dynamic curriculum via Potential-Aware Query Mining (Phase 0, \S\ref{sec:paqm}); second, we stabilize the optimization process using Hybrid Stratified Replay, which combines diagnostic stratification (Phase 1, \S\ref{sec:stratification}) with structured group assembly (Phase 2, \S\ref{sec:assembly}). By addressing both data ingestion and optimization-group composition, our framework targets zero-signal groups and premature entropy collapse.

\subsection{Phase 0: Potential-Aware Query Mining (PAQM)}
\label{sec:paqm}

Constructing a high-quality prompt set is the first line of defense against training instability. We introduce Potential-Aware Query Mining (PAQM), a dynamic curation pipeline that filters queries based on their "Learnability."

\paragraph{Mathematical Definition of Learnability.}
For a given prompt $x$, we sample $N=16$ rollouts $\{y_1, ..., y_{16}\}$ from the current policy $\pi_\theta$. We estimate the empirical success rate $\hat{p}(x) = \frac{1}{N} \sum r(y_i)$.
We categorize queries into three zones:
\begin{itemize}
    \item \textbf{Distillation Zone (Learnable):} Queries where $0 < \hat{p}(x) < 1$. These exhibit variance in outcomes, indicating the model is capable but unstable---the ideal target for RL.
    \item \textbf{Trivial Zone:} Queries where $\hat{p}(x) = 1$. These are mastered concepts and are down-sampled to prevent overfitting.
    \item \textbf{Impossible Zone:} Queries where $\hat{p}(x) = 0$. These yield zero gradients in standard RL and are candidates for diagnosis.
\end{itemize}

To obtain a tractable rollout-level confidence score within the Distillation Zone, we define Path Entropy as the length-normalized sequence surprisal of a sampled rollout. Motivated by predictive-uncertainty estimation \citep{malinin2018predictive}, we compute
\begin{equation}
    H_{\mathrm{path}}(y|x) = -\frac{1}{|y|} \sum_{t=1}^{|y|} \log \pi_\theta(y_t | x, y_{<t}).
\end{equation}
This sample-level quantity is a proxy for model confidence: it is not the Shannon entropy of the full predictive distribution and should not be interpreted as a direct estimate of epistemic uncertainty.

\paragraph{Adaptive Frontier Expansion via Modality-Decoupled Diagnosis.}
For queries in the Impossible Zone, we apply Modality-Decoupled Diagnosis. We probe the model using ground-truth captions to isolate the failure mode.
\begin{itemize}
    \item \textbf{Perceptual Deficit:} If the model solves the text-only probe, we inject a Visual Decomposition Hint (e.g., coordinates, shapes) derived from an auxiliary vision model.
    \item \textbf{Cognitive Deficit:} If the model fails the probe, we inject a Key Reasoning Node (e.g., the first theorem step) distilled from a teacher model's trace.
\end{itemize}
This mechanism reclaims roughly 30\% of hard queries back into the Distillation Zone. Note that these hints are generated offline using the proprietary teacher model Seed-1.5-VL \citep{guo2025seed} and are injected only during Phase 0 curriculum construction.

\subsection{Phase 1: Diagnostic Path-Entropy--Reward Stratification}
\label{sec:stratification}

One source of instability in Group Relative Policy Optimization (GRPO) is the stochastic quality of optimization groups. To address it, we propose Hybrid Stratified Replay (HSR), which treats "Mastered Samples" and "Confident Errors" as Stability Anchors.

We categorize rollouts based on outcome correctness $r(y)$ and the Path-Entropy score $H_{\mathrm{path}}(y|x)$. To adapt to the shifting policy distribution, we employ Dynamic Percentile Thresholding. We calculate the empirical Path-Entropy distribution $\mathcal{H}$ across all rollouts in the current global batch (covering multiple queries) to obtain batch-adaptive cutoffs. We define dynamic thresholds $\tau_{low} = \text{Percentile}(\mathcal{H}, 10)$ and $\tau_{high} = \text{Percentile}(\mathcal{H}, 90)$.
Rollouts are classified into three buffers:
\begin{itemize}
    \item \textbf{Stability Anchors} ($\mathcal{B}_{anc}$): Correct samples with low Path Entropy ($r=1, H_{\mathrm{path}}(y|x) < \tau_{low}$). These are reused as high-confidence positive anchors.
    \item \textbf{Hard Negatives} ($\mathcal{B}_{neg}$): Incorrect samples with low Path Entropy ($r=0, H_{\mathrm{path}}(y|x) < \tau_{low}$). These are high-confidence errors included to provide outcome contrast.
    \item \textbf{Epiphanies} ($\mathcal{B}_{epi}$): Correct samples with high Path Entropy ($r=1, H_{\mathrm{path}}(y|x) > \tau_{high}$). These are low-confidence successes emphasized as learning targets.
\end{itemize}
Samples in the "Confusion Zone" ($r=0, H_{\mathrm{path}}(y|x) > \tau_{high}$) are treated as low-confidence errors and excluded from the valid sample pool.

\subsection{Phase 2: Hybrid Group Assembly}
\label{sec:assembly}

Standard GRPO constructs optimization groups via random sampling, which can yield low-contrast or high-variance updates \citep{shuffleR1_2025,leng2025mmr1}. To balance the need for \textit{targeted correction} and \textit{natural on-policy coverage}, we allocate the total group budget $G$ into two equal streams: the Structured Stream ($50\%$) and the Natural Stream ($50\%$).

\paragraph{Stream A: Structured Stabilization (50\%).}
This stream focuses on "Correction" by explicitly engineering the batch topology. Within the allocated budget for this stream ($G_{struct} = 0.5G$), we enforce a strict semantic ratio: 50\% Epiphanies (to drive learning), 25\% Stability Anchors (to anchor the baseline), and 25\% Hard Negatives (to sharpen the decision boundary).
Formally:
\begin{equation}
\begin{split}
    G_{struct} = & \, S_{epi}^{50\%} \cup \text{Sample}(\mathcal{B}_{anc}, 25\%) \\
                 & \cup \text{Sample}(\mathcal{B}_{neg}, 25\%)
\end{split}
\label{eq:hybrid_assembly}
\end{equation}
When all structured quotas are available, this injection grounds the selected exploration targets with positive anchors and contrasting errors within the micro-batch.

\paragraph{Stream B: Natural Generalization (50\%).}
To prevent the policy from overfitting to the artificial topology of Stream A, the second stream ($G_{nat} = 0.5G$) focuses on "Robustness." We sample groups uniformly from the valid sample pool (excluding only the filtered noise). This preserves natural on-policy coverage alongside the deliberately shifted structured stream.

\paragraph{Clarification on Replay Scope (Intra-Step Only).}
A critical distinction of our framework is the scope of the replay buffers. Unlike traditional experience replay used in off-policy RL, HSR employs an Intra-Step Replay mechanism. The buffers $\mathcal{B}_{anc}$ and $\mathcal{B}_{neg}$ have a lifespan of $\tau=1$; they are populated solely from the $N$ rollouts generated at the \textit{current} optimization step and are cleared immediately after the gradient computation. This ensures that all "reused" samples are drawn from the current policy $\pi_\theta$, thereby maintaining on-policy consistency without requiring importance sampling corrections for stale data.

\subsubsection{Theoretical Motivation: Degenerate Groups}
\label{sec:theory}

Randomly sampled groups may become reward-degenerate when all rollouts receive the same outcome, causing group-relative advantages to vanish. HSR reduces the incidence of such zero-signal groups by constructing a structured stream containing both successful and failed rollouts whenever the corresponding samples are available. Section~\ref{sec:appendix_theory_detailed} provides a conditional variance analysis and describes the fallback used when a structured quota cannot be satisfied.

\section{Experiments}

\subsection{Experimental Setup}

\paragraph{Datasets and Benchmarks.}
Distinct from prior multi-stage approaches, Stable-MM-R1 uses RL-only parameter updates on multimodal data, without a supervised cold-start phase; teacher-generated hints affect selected training prompts but are not used for supervised parameter updates.
To support this, we construct a high-quality RL Prompt Corpus comprising approximately 189k samples by aggregating verifiable multimodal tasks from Geometry3K \citep{lu2021intergps} and a curated subset of MM-Eureka \citep{meng2025mmeureka}. Furthermore, we integrate ViRL39K \citep{wang2025vlrethinker}, a large-scale dataset of verifiable visual reasoning queries, to expand the diversity of visual scenarios.
Crucially, our \textit{Potential-Aware Query Mining (PAQM)} pipeline utilizes a teacher-guided distillation strategy to generate \textit{scaffolding hints} for hard queries. Given the multimodal nature of our data source, we exclusively employ the proprietary Vision-Language Model Seed-1.5-VL \citep{guo2025seed} to generate both perceptual descriptions and reasoning traces offline. These cached multimodal hints are injected into selected prompts during Phase 0 curriculum construction, before RL rollouts begin.

For evaluation, we employ a comprehensive suite of benchmarks covering mathematical reasoning, visual perception, and chart understanding: MathVerse \citep{zhang2024mathverse}, MathVision \citep{wang2024mathvision}, MathVista \citep{lu2023mathvista}, DynaMath \citep{zou2024dynamath}, WeMath \citep{qiao2024wemath}, LogicVista \citep{xiao2024logicvista}, MMMU \citep{yue2024mmmu}, MMMU-Pro \citep{yue2024mmmupro}, CharXiv \citep{wang2024charxiv}, and MMStar \citep{chen2024mmstar}.

\paragraph{Baselines.}
We benchmark Stable-MM-R1 against three categories of systems to ensure a rigorous evaluation.
First, we include leading proprietary models as upper-bound references: GPT-4.1 \citep{openai2025gpt41}, Gemini-2.5-Pro-Thinking \citep{comanici2025gemini}, Claude-4-Sonnet \citep{anthropic2025claude}, and Doubao-1.5-Thinking-Vision-Pro \citep{guo2025seed}.
Second, we consider representative lightweight open-source MLLMs: Qwen2.5-VL-3B-Instruct \citep{bai2025qwen25vl}, FAST-3B \citep{xiao2025fast}, and VLAA-Thinker-3B \citep{chen2025vlaa}.
Third, we evaluate competitive 7B-scale open-source MLLMs, including Qwen2.5-VL-7B-Instruct, OpenVLThinker-1.2-7B \citep{deng2025openvlthinker}, MM-Eureka-Qwen-7B \citep{meng2025mmeureka}, MMR1-Math-v0 \citep{leng2025mmr1}, ThinkLite-7B-VL \citep{wang2025thinklite}, VLAA-Thinker-7B, VL-Rethinker-7B \citep{wang2025vlrethinker}, and Vision-G1 \citep{zha2025visiong1}.
Most open-source baselines are fine-tuned from the Qwen2.5-VL family, ensuring comparability in architecture and training setup.

\paragraph{Implementation Details.}
We utilize the EasyR1 framework \citep{yaowei2025easyr1} as our training codebase. We employ Qwen2.5-VL-7B-Instruct \citep{bai2025qwen25vl} as the base policy to verify scalability.
We freeze the vision encoder and update only the LLM backbone during RL training.
We set the global batch size to 128 and the learning rate to $1\times10^{-6}$. The rollout temperature is set to 1.0.
For HSR, we generate $N=16$ rollouts per query. Across in-house methods, we match the final group size ($G=16$) and the number of gradient updates (400 steps). This controls the optimization-update budget but does not make total data-preparation and generation cost identical: PAQM uses an offline hint cache and may resample queries that initially fall in the Impossible Zone. All experiments are conducted on $8\times$ H100 GPUs.
Throughout the paper, Pass@8 denotes majority-vote accuracy over eight independently sampled responses.

\paragraph{Data Decontamination.}
Given the aggregation of multiple datasets, data leakage is a critical concern. We applied a de-duplication pipeline. For text, we used MinHash LSH to remove samples with $>0.8$ Jaccard similarity to any query in the evaluation set. For images, we employed perceptual hashing (pHash) to filter out visual duplicates. These checks reduce direct and near-duplicate overlap, although they cannot rule out all semantic contamination.


\paragraph{Controlled Baseline Reproduction \& Fairness.}
All in-house comparisons use the same base model, source prompt pool, final per-query rollout group size, training schedule, and evaluation protocol. PAQM and HSR are enabled only in the configurations explicitly marked in Table~\ref{tab:ablation}. Standard GRPO and DAPO do not use PAQM. The `+ PAQM` configuration uses the full hint-augmented curriculum with the standard GRPO objective, whereas Full Stable-MM-R1 additionally enables HSR; their comparison therefore isolates the contribution of structured batch topology under the same PAQM data pipeline. We match the final group size ($G=16$) and update count across methods; the offline teacher and resampling overhead described above are additional costs of PAQM.

\textit{Note on Training Duration:} While Stable-MM-R1 maintained stability throughout the full training schedule (approx. 400 steps), we observed that the GRPO and DAPO baselines frequently suffered from entropy collapse and performance degradation after step 250 (as visualized in Figure \ref{fig:training_dynamics_stacked}). To ensure a fair comparison, we report the peak performance achieved by these baselines prior to collapse, rather than their final converged checkpoints. This protocol ensures that the reported gaps reflect the ceiling of the baselines capabilities rather than their failure modes.
\subsection{Main Results}


\begin{table*}[t!]
\centering
\small
\setlength{\tabcolsep}{1.8pt} 
\renewcommand{\arraystretch}{1.25} 

\resizebox{\linewidth}{!}{
\begin{tabular}{lcccccccccc|r}
\toprule
\rowcolor{white}
\textbf{Model} &
\textbf{MathVerse-V} & \textbf{MathVision} & \textbf{MathVista} &
\textbf{DynaMath-W} & \textbf{WeMath} & \textbf{LogicVista} &
\textbf{MMMU} & \textbf{MMMU-Pro} & \textbf{CharXiv} & \textbf{MMStar}&
\textbf{Avg.} \\
\midrule

\rowcolor{graybg}
\multicolumn{12}{l}{\textit{\textbf{Closed-Source Models}}} \\
Gemini-2.5-Pro-Thinking & \textit{81.2} & \textit{55.3} & \textit{83.8} & \textit{57.1} & 78.0 & \textit{75.2} & 82.0 & \textit{76.5} & \textit{69.3} & 79.7& 73.8\\
Doubao-1.5-thinking-vision-pro & \textit{80.4} & 68.7 & 85.6 & \textit{60.5} & \textit{78.0} & \textit{71.8} & 77.9 & 67.6 & \textit{63.4} & 78.2& 73.2\\
Claude-4-Sonnet & \textit{66.1} & \textit{54.6} & \textit{70.4} & \textit{46.9} & \textit{63.0} & \textit{64.4} & 74.4 & \textit{60.7} & \textit{58.4} & 66.9 & 62.6\\
GPT-4.1 & \textit{59.8} & \textit{51.8} & 72.0 & \textit{48.3} & 55.5 & \textit{63.8} & 75.0 & \textit{65.0} & \textit{55.8} & 71.2 & 61.8\\
\midrule

\rowcolor{graybg}
\multicolumn{12}{l}{\textit{\textbf{3B-scale MLLMs}}} \\
Qwen2.5-VL-3B-Instruct & \textit{33.0} & 21.2 & 62.3 & \textit{17.0} & \textit{17.9} & 35.8 & \underline{53.1} & 31.6 & \textit{25.3} & 51.1 & 34.8\\
VLAA-Thinker-3B & 36.4 & 24.4 & 61.0 & \underline{18.2} & \underline{33.8} & \underline{38.5} & \textit{49.7} & \textit{33.3} & \textit{28.2} & 53.4 & \underline{37.7}\\
FAST-3B & \underline{\textit{37.2}} & \underline{26.8} & \underline{66.2} & \textit{15.4} & \textit{23.3} & \textit{35.4} & \textit{52.0} & \underline{\textit{34.6}} & \underline{\textit{28.3}} & \underline{55.2} & 37.4\\
\rowcolor{lightblue}
\textbf{Stable-MM-R1-3B} & \textbf{47.9} & \textbf{44.9} & \textbf{67.2} & \textbf{24.4} & \textbf{41.9} & \textbf{44.1} & \textbf{56.5} & \textbf{43.9} & \textbf{34.4} & \textbf{57.2} & \textbf{46.2}\\
\textit{$\Delta$ (Ours--Best Evaluated Open 3B)} & \gbf{+10.7} & \gbf{+18.1} & \gbf{+1.0} & \gbf{+6.2} & \gbf{+8.1} & \gbf{+5.6} & \gbf{+3.4} & \gbf{+9.3} & \gbf{+6.1} & \gbf{+2.0} & \gbf{+8.5} \\
\midrule

\rowcolor{graybg}
\multicolumn{12}{l}{\textit{\textbf{7B-scale Models}}} \\
Qwen2.5-VL-7B-Instruct & \textit{42.9} & 25.1 & 68.2 & \textit{21.2} & 36.2 & \textit{45.0} & 58.6 & 38.3 & \textit{35.5} & 62.1 & 43.3\\
VLAA-Thinker-7B & 48.2 & 26.4 & 68.0 & 22.4 & \textit{29.2} & 48.5 & \textit{54.6} & \textit{41.6} & \textit{36.1} & \underline{64.5} & 44.0\\
ThinkLite-7B-VL & \textit{45.3} & \underline{32.9} & 75.1 & \textit{22.0} & \textit{26.5} & \textit{40.7} & 55.5 & \textit{41.3} & \textit{39.3} & 63.4 & 44.2\\
OpenVLThinker-1.2-7B & \textit{40.7} & 25.9 & 72.3 & \textit{21.2} & \textit{37.9} & \textit{41.4} & \underline{58.7} & 42.9 & \textit{39.3} & 62.9 & 44.3\\
VL-Rethinker-7B & \textit{49.1} & 32.3 & \underline{74.9} & \underline{27.4} & \textit{27.8} & \textit{44.5} & 56.7 & 41.7 & \textit{39.8} & 62.7 & 45.7\\
MMR1-Math-v0 & 45.1 & 30.2 & 71.0 & \textit{25.2} & \textit{33.2} & \underline{50.8} & \textit{57.1} & \textit{43.2} & \textit{39.3} & 63.8 & 45.9\\
MM-Eureka-Qwen-7B & \textit{49.6} & 26.9 & 73.0 & \textit{24.0} & 34.7 & \textit{46.8} & \textit{57.3} & \underline{\textit{43.3}} & \textit{39.5} & 64.4 & 46.0\\
Vision-G1 & \underline{\textit{50.0}} & 31.3 & \textbf{76.1} & \textit{27.2} & \underline{29.0} & 50.2 & 53.4 & 41.2 & \underline{\textit{41.0}} & \textbf{66.0} & \underline{46.5}\\
\rowcolor{lightblue}
\textbf{Stable-MM-R1-7B} & \textbf{56.3} & \textbf{53.1} & \underline{75.5} & \textbf{39.3} & \textbf{49.6} & \textbf{55.3} & \textbf{67.2} & \textbf{52.3} & \textbf{47.7} & \underline{65.9} & \textbf{56.2}\\
\textit{$\Delta$ (Ours--Best Evaluated Open 7B)} & \gbf{+6.3} & \gbf{+20.2} & \rtext{-0.6} & \gbf{+11.9} & \gbf{+11.7} & \gbf{+4.5} & \gbf{+8.5} & \gbf{+9.0} & \gbf{+6.7} & \rtext{-0.1} & \gbf{+9.7} \\
\bottomrule
\end{tabular}
}
\caption{\footnotesize Performance comparison of various MLLMs on diverse multimodal reasoning benchmarks. \textbf{Note on Metrics:} We report Pass@8 (majority-vote) accuracy for all evaluated open-source models. Proprietary models are included as reference points and excluded from the within-scale rankings. Within each open-source model group (3B and 7B), the best results are highlighted in bold, and the second-best are underlined. MathVerse-V, DynaMath-W and WeMath-S denote the vision-only, worst, and strict settings, respectively.}
\label{tab:multimodal_performance_revised}
\end{table*}

\begin{figure}[t]
    \centering
    \begin{subfigure}[b]{\columnwidth}
        \centering
        \includegraphics[width=\linewidth]{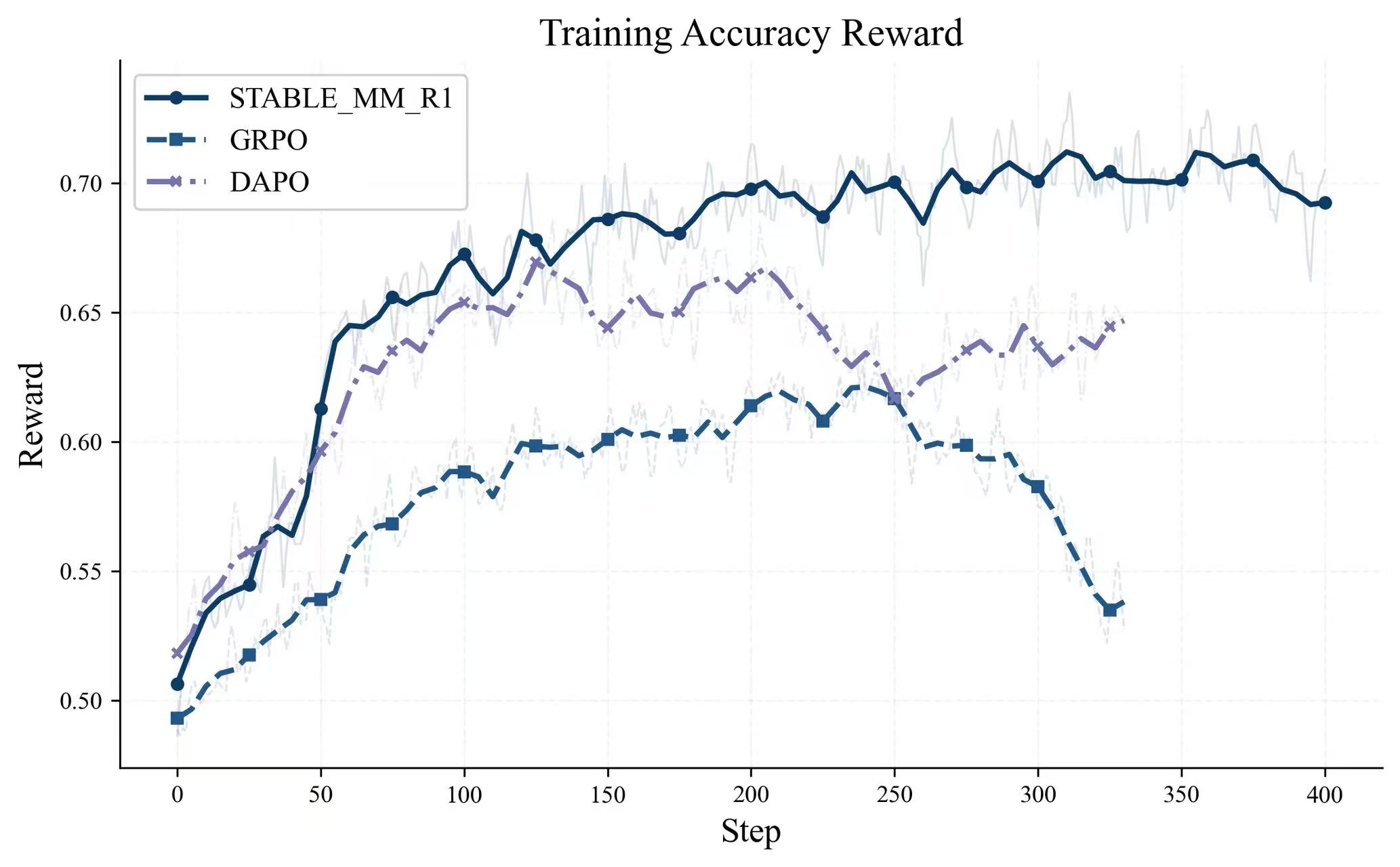}
        \caption{Accuracy dynamics.}
        \label{fig:reward_dynamics}
    \end{subfigure}

    \vspace{2mm}

    \begin{subfigure}[b]{\columnwidth}
        \centering
        \includegraphics[width=\linewidth]{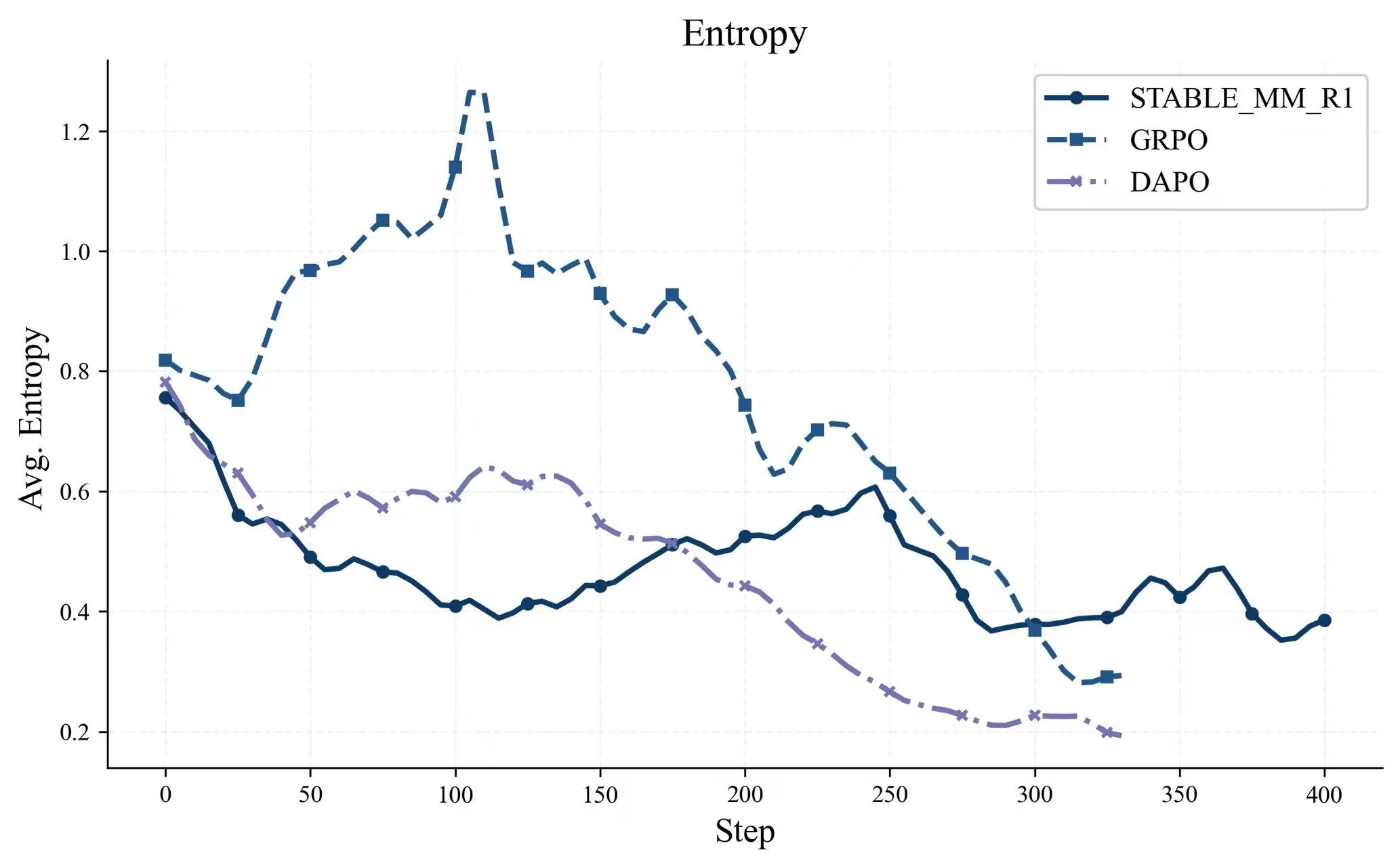}
        \caption{Path-Entropy evolution.}
        \label{fig:entropy_dynamics}
    \end{subfigure}

    \caption{\textbf{Training stability analysis.} Accuracy (top) and mean Path Entropy (bottom) over training. Stable-MM-R1 maintains healthy exploration levels and avoids the collapse modes of the baselines.}
    \label{fig:training_stability}
    \vspace{-0.2in}
\end{figure}

\paragraph{Comparison with Baselines.}
Table \ref{tab:multimodal_performance_revised} presents a comprehensive comparison against both proprietary and open-source models.
Stable-MM-R1 achieves the highest average score among the evaluated open-source 7B-scale models, with an average score of 56.2\%, outperforming the strongest evaluated open-source competitor Vision-G1 (46.5\%) by 9.7 points.
Notably, on the fine-grained visual reasoning benchmark MathVerse, our method achieves 56.3\%, surpassing Vision-G1 (50.0\%) by 6.3 points and indicating strong performance on fine-grained visual mathematical reasoning.
On MathVision, which evaluates challenging visual mathematical reasoning, we observe an improvement of 28.0 points over the Qwen2.5-VL base model and 20.2 points over the strongest evaluated open-source baseline.
While our performance on MathVista (75.5\%) is slightly below Vision-G1 (-0.6\%), Stable-MM-R1 outperforms the base model and the other evaluated RL baselines on average while exhibiting more stable training dynamics in Figures~\ref{fig:training_stability} and~\ref{fig:training_dynamics_stacked}.
\paragraph{Stability Analysis.}
Figure~\ref{fig:training_stability} contrasts the training dynamics of the three methods. The GRPO baseline exhibits large Path-Entropy fluctuations followed by a sharp decline, while DAPO's Path Entropy gradually falls to approximately $0.2$ by its final recorded step. After its initial decline, Stable-MM-R1 remains mostly within approximately $0.35$--$0.60$ nats, and its reward stays near the late-training plateau through step 400. These trajectories show greater late-training stability for Stable-MM-R1 in this run, without implying monotonic improvement at every step.

\begin{table}[t!]
\centering
\renewcommand{\arraystretch}{1.2}
\resizebox{\linewidth}{!}{%
\begin{tabular}{lcccc}
\toprule
\textbf{Configuration} & \textbf{MathVista} & \textbf{MathVision} & \textbf{MMMU} & \textbf{Avg.} \\
\midrule
GRPO (Baseline) & 70.6 & 50.1 & 61.5 & 60.7 \\
DAPO (Strong Baseline) & 71.8 {\color{graygain}(+1.2)} & 51.2 {\color{graygain}(+1.1)} & 62.9 {\color{graygain}(+1.4)} & 62.0 {\color{graygain}(+1.3)} \\
\midrule
+ PAQM (Ours) & 73.5 {\color{deeppink}(+2.9)} & 52.9 {\color{deeppink}(+2.8)} & 65.1 {\color{deeppink}(+3.6)} & 63.8 {\color{deeppink}(+3.1)} \\
+ HSR (Ours) & \underline{74.2} {\color{deeppink}(+3.6)} & \underline{53.6} {\color{deeppink}(+3.5)} & \underline{66.2} {\color{deeppink}(+4.7)} & \underline{64.7} {\color{deeppink}(+4.0)} \\
\textbf{Full Stable-MM-R1} & \textbf{75.5} {\color{deeppink}(+4.9)} & \textbf{54.7} {\color{deeppink}(+4.6)} & \textbf{67.2} {\color{deeppink}(+5.7)} & \textbf{65.8} {\color{deeppink}(+5.1)} \\
\bottomrule
\end{tabular}%
}
\caption{\footnotesize \textbf{Ablation study.} Standard GRPO and DAPO use neither PAQM nor HSR. `+ PAQM` applies the full hint-augmented curriculum with the standard GRPO objective, `+ HSR` enables HSR without PAQM, and Full Stable-MM-R1 combines both components. The comparison between `+ PAQM` and Full isolates the contribution of HSR. Best results are \textbf{bold}.}
\label{tab:ablation}
\end{table}

    


\subsection{Ablation Study}

To dissect the contribution of each component, we analyze the impact of removing PAQM and HSR. Table \ref{tab:ablation} summarizes the results.

\paragraph{HSR vs. Hints.}
A key question is whether the gains stem solely from the Seed-1.5-VL hints in PAQM. The row `+ PAQM` represents GRPO trained on the same hint-augmented curriculum used by the full method. Comparing `+ PAQM` with `Full Stable-MM-R1` therefore isolates a 2.0-point average gain from adding HSR under PAQM, showing that structured batch topology contributes beyond the hint-augmented curriculum alone.

\paragraph{Effectiveness of PAQM.}
As shown in Table \ref{tab:ablation}, adding PAQM alone (`+ PAQM`) improves the average score by 3.1 points over the GRPO baseline. This result is consistent with the benefit of concentrating updates on the learnable frontier.

\begin{figure}[t]
    \centering
    \begin{subfigure}[b]{\columnwidth}
        \centering
        \includegraphics[width=\linewidth]{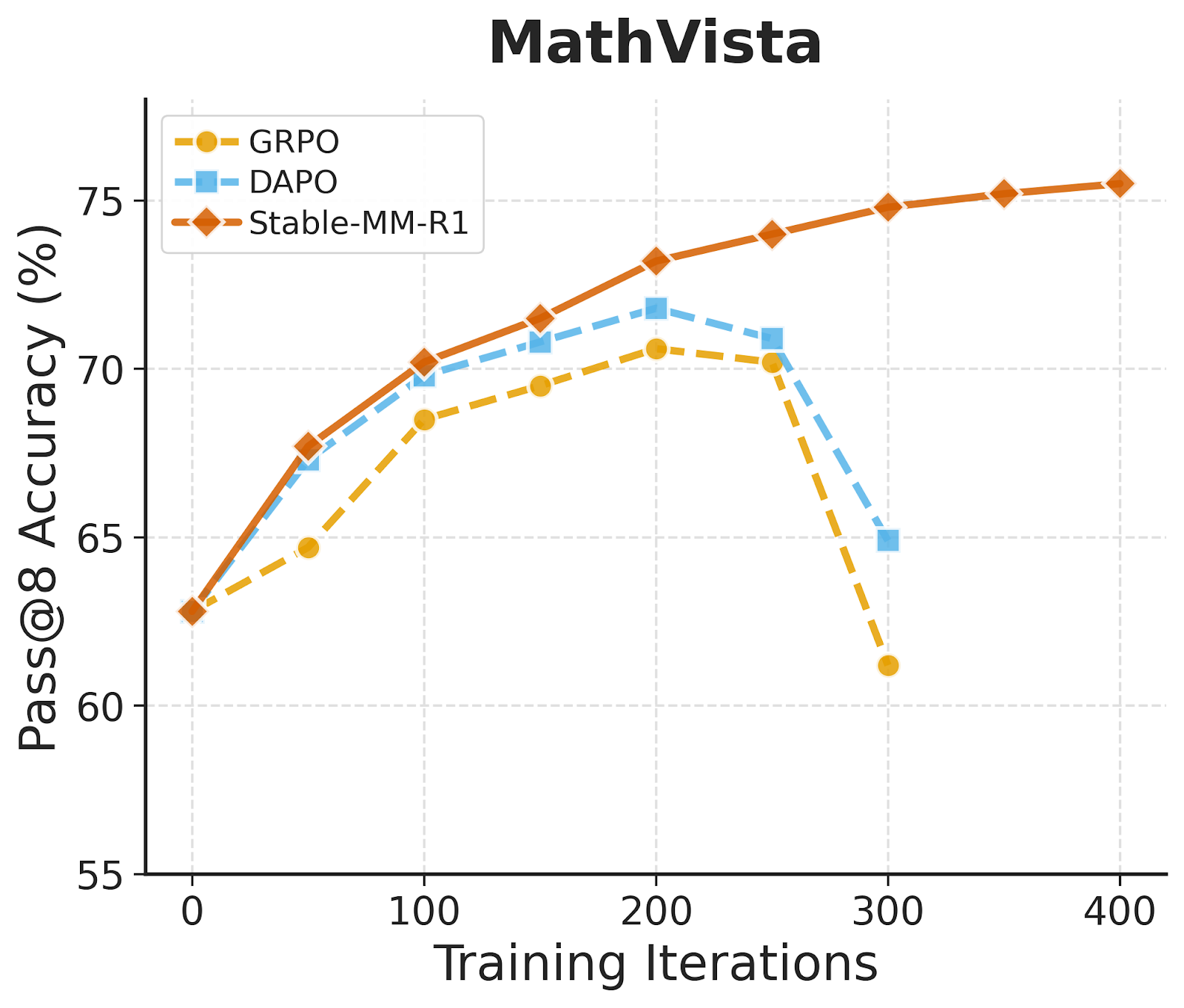}
        \caption{MathVista Pass@8 over training.}
        \label{fig:mathvista_dynamics}
    \end{subfigure}

    \vspace{3mm}

    \begin{subfigure}[b]{\columnwidth}
        \centering
        \includegraphics[width=\linewidth]{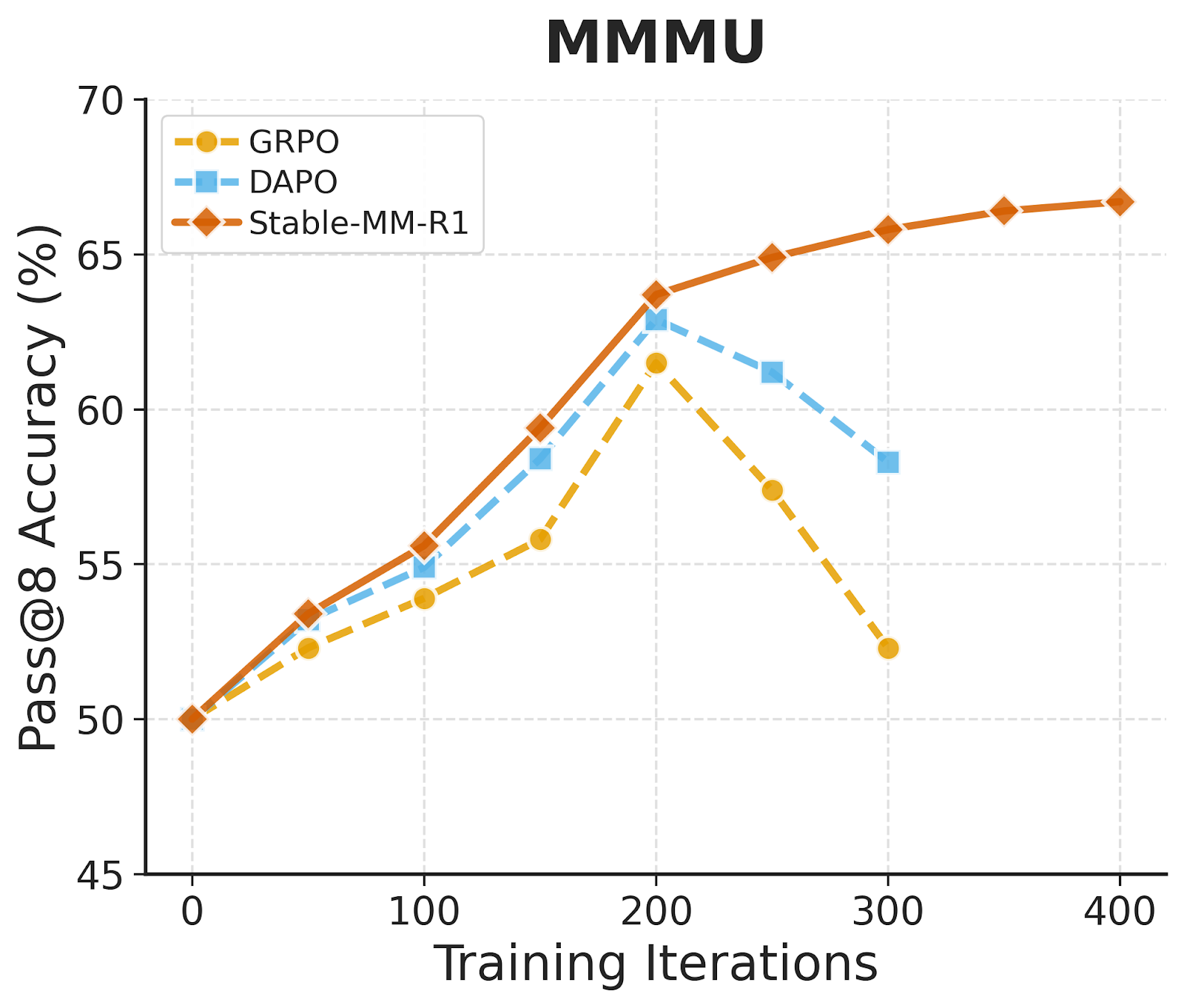}
        \caption{MMMU Pass@8 over training.}
        \label{fig:mmmu_dynamics}
    \end{subfigure}

    \caption{\textbf{Training dynamics comparison.} Pass@8 accuracy trajectories on (a) MathVista and (b) MMMU. While GRPO and DAPO suffer from performance collapse, Stable-MM-R1 maintains a robust upward trajectory.}
    \label{fig:training_dynamics_stacked}
\end{figure}

\paragraph{Impact of Hybrid Stratified Replay.}
Adding HSR without PAQM (`+ HSR`) improves the average score by 4.0 points over GRPO, while combining PAQM and HSR yields the best ablation result (65.8 average). Figure~\ref{fig:training_dynamics_stacked} further shows that the full method continues to improve after the GRPO and DAPO curves peak; because the figure compares complete methods rather than individual HSR buffer components, we do not attribute this trend to any single buffer category.


\section{Conclusion}

In this paper, we showed that stochastic optimization-group composition can contribute to instability in RL training for reasoning. We proposed a unified data-centric framework comprising Potential-Aware Query Mining (PAQM) and Hybrid Stratified Replay (HSR). By dynamically curating learnable queries and structuring optimization groups with stability anchors and contrastive negatives, the full method mitigated entropy collapse in our experiments.

Beyond the performance gains, our results show that explicit contrast within the micro-batch---via anchors and hard negatives---can improve long-chain reasoning training without changing the underlying GRPO objective. This data-centric paradigm offers a complementary path for stabilizing future multimodal models.

\section*{Limitations}

While Stable-MM-R1 improves stability, it relies on a proprietary teacher model (Seed-1.5-VL) for initial scaffolding, which may introduce distillation bias and accessibility constraints. Although our comparisons match update count and final group size, PAQM incurs additional offline teacher-generation and query-resampling costs, and HSR requires careful intra-step buffer management. Finally, our method currently focuses on verifiable reasoning; extending entropy-guided stratification to open-ended generation remains future work.
\section*{Ethical Considerations}

As this work introduces a framework for enhancing the reasoning capabilities of Multimodal Large Language Models (MLLMs) via Reinforcement Learning, we address several ethical implications regarding data selection, model behavior, and potential misuse.

\paragraph{Bias in Data Selection (PAQM).}
Our Potential-Aware Query Mining (PAQM) mechanism actively filters training data based on the model's current capability (Pass@K). While this improves efficiency, there is a theoretical risk of \textit{algorithmic exclusion}. If the base model exhibits pre-existing biases against specific linguistic dialects, cultural contexts, or visual concepts (classifying them as "Impossible"), PAQM might systematically exclude this data from the training loop, exacerbating the bias. We mitigate this by using diverse, standard academic benchmarks (STEM-focused), but we advise practitioners to audit the "Discarded" data partition when applying this method to sociologically sensitive domains.

\paragraph{Data Privacy and Content Safety.}
We inspected the training corpus to identify and mitigate potential risks related to Personally Identifiable Information (PII) and offensive content. Although our data sources (Geometry3K, MM-Eureka, and ViRL39K) primarily consist of public academic benchmarks in STEM domains where PII risks are relatively low, we employed automated keyword filtering and random-sample audits to reduce the risk of propagating sensitive personal data or toxic material into the RL fine-tuning stage.

\paragraph{Teacher-Student Alignment Risks.}
The use of teacher-generated hints (visual decompositions and reasoning nodes) facilitates rapid learning. However, this creates a dependency on the safety alignment of the teacher model. If the teacher model harbors latent jailbreak vulnerabilities or hallucinates harmful instructions, the student model may internalize these patterns via the high-reward gradient signal. Although our experiments focus on verifiable math and logic tasks where safety risks are minimal, applying Stable-MM-R1 to open-ended generation requires integrating safety-specific reward models (e.g., Rule-Based Reward Models or Constitutional AI) to prevent the reinforcement of hazardous content.

\paragraph{Dual Use and Automation.}
By stabilizing the reasoning capabilities of 7B-scale models, this work lowers the barrier for deploying powerful multimodal agents. While this democratizes access to advanced AI, it also increases the potential for misuse in automated tasks, such as solving CAPTCHAs or automating interaction in visual environments. We emphasize that the models trained in this paper are research prototypes intended for academic evaluation and should be subjected to rigorous red-teaming before real-world deployment.

\paragraph{Environmental Impact.}
RL fine-tuning is computationally intensive. Our experiments utilized $8\times$ H100 GPUs. HSR is designed to reduce zero-signal optimization groups under a fixed step budget, although we do not report a full energy or carbon accounting. Data-centric efficiency remains an important direction for reducing the environmental cost of model post-training.

\paragraph{Disclosure of AI Assistance.}
We acknowledge the use of AI assistants for improving textual clarity and verifying code syntax. The authors reviewed the manuscript, scientific claims, and reported results and retain responsibility for the final content.

\bibliography{custom}

@article{llavaonevision,
  title={{LLaVA-OneVision-1.5}: Fully Open Framework for Democratized Multimodal Training},
  author={An, Xiang and Xie, Yin and Yang, Kaicheng and Zhang, Wenkang and Zhao, Xiuwei and Cheng, Zheng and Wang, Yirui and Xu, Songcen and Chen, Changrui and Zhu, Didi and Wu, Chunsheng and Tan, Huajie and Li, Chunyuan and Yang, Jing and Yu, Jie and Wang, Xiyao and Qin, Bin and Wang, Yumeng and Yan, Zizhen and Feng, Ziyong and Liu, Ziwei and Li, Bo and Deng, Jiankang},
  journal={arXiv preprint arXiv:2509.23661},
  year={2025}
}

@article{tang2025rethinking,
  author       = {Xinyu Tang and Yuliang Zhan and Zhixun Li and Wayne Xin Zhao and Zhenduo Zhang and Zujie Wen and Zhiqiang Zhang and Jun Zhou},
  title        = {Rethinking Sample Polarity in Reinforcement Learning with Verifiable Rewards},
  journal={arXiv preprint arXiv:2512.21625},
  year         = {2025},
}

@misc{deepseekr1_2025,
  title        = {DeepSeek-R1: Incentivizing Reasoning Capability in LLMs via Reinforcement Learning},
  author       = {DeepSeek-AI and Guo, Daya and Yang, Dejian and Zhang, Haowei and Song, Junxiao and Zhang, Ruoyu and Xu, Runxin and Zhu, Qihao and Ma, Shirong and Wang, Peiyi and others},
  year         = {2025},
  eprint       = {2501.12948},
  archivePrefix= {arXiv},
  primaryClass = {cs.CL},
  doi          = {10.48550/arXiv.2501.12948}
}

@misc{shao2024deepseekmath,
  title        = {DeepSeekMath: Pushing the Limits of Mathematical Reasoning in Open Language Models},
  author       = {Shao, Zhihong and Wang, Peiyi and Zhu, Qihao and Xu, Runxin and Song, Junxiao and Bi, Xiao and Zhang, Haowei and Zhang, Mingchuan and Li, Y. K. and Wu, Y. and Guo, Daya},
  year         = {2024},
  eprint       = {2402.03300},
  archivePrefix= {arXiv},
  primaryClass = {cs.CL},
  doi          = {10.48550/arXiv.2402.03300}
}

@misc{shuffleR1_2025,
  title        = {Shuffle-R1: Efficient RL framework for Multimodal Large Language Models via Data-centric Dynamic Shuffle},
  author       = {Zhu, Linghao and Guan, Yiran and Liang, Dingkang and Ju, Jianzhong and Luo, Zhenbo and Qin, Bin and Luan, Jian and Liu, Yuliang and Bai, Xiang},
  year         = {2025},
  eprint       = {2508.05612},
  archivePrefix= {arXiv},
  primaryClass = {cs.CV},
  doi          = {10.48550/arXiv.2508.05612}
}

@misc{dapoyuscale_2025,
  title        = {DAPO: An Open-Source LLM Reinforcement Learning System at Scale},
  author       = {Yu, Qiying and Zhang, Zheng and Zhu, Ruofei and Yuan, Yufeng and Zuo, Xiaochen and Yue, Yu and Dai, Weinan and Fan, Tiantian and Liu, Gaohong and Liu, Lingjun and Liu, Xin and Lin, Haibin and Lin, Zhiqi and Ma, Bole and Sheng, Guangming and Tong, Yuxuan and Zhang, Chi and Zhang, Mofan and Zhang, Wang and Zhu, Hang and Zhu, Jinhua and Chen, Jiaze and Chen, Jiangjie and Wang, Chengyi and Yu, Hongli and Song, Yuxuan and Wei, Xiangpeng and Zhou, Hao and Liu, Jingjing and Ma, Wei-Ying and Zhang, Ya-Qin and Yan, Lin and Qiao, Mu and Wu, Yonghui and Wang, Mingxuan},
  year         = {2025},
  eprint       = {2503.14476},
  archivePrefix= {arXiv},
  primaryClass = {cs.LG},
  doi          = {10.48550/arXiv.2503.14476}
}

@misc{yuan2025ppocollapse,
  title        = {What's Behind PPO's Collapse in Long-CoT? Value Optimization Holds the Secret},
  author       = {Yuan, Yufeng and Yue, Yu and Zhu, Ruofei and Fan, Tiantian and Yan, Lin},
  year         = {2025},
  eprint       = {2503.01491},
  archivePrefix= {arXiv},
  primaryClass = {cs.LG},
  doi          = {10.48550/arXiv.2503.01491}
}

@misc{schulman2017ppo,
  title        = {Proximal Policy Optimization Algorithms},
  author       = {Schulman, John and Wolski, Filip and Dhariwal, Prafulla and Radford, Alec and Klimov, Oleg},
  year         = {2017},
  eprint       = {1707.06347},
  archivePrefix= {arXiv},
  primaryClass = {cs.LG},
  doi          = {10.48550/arXiv.1707.06347}
}

@misc{ouyang2022instructgpt,
  title        = {Training language models to follow instructions with human feedback},
  author       = {Ouyang, Long and Wu, Jeffrey and Jiang, Xu and Almeida, Diogo and Wainwright, Carroll L. and Mishkin, Pamela and Zhang, Chong and Agarwal, Sandhini and Slama, Katarina and Ray, Alex and Schulman, John and Hilton, Jacob and Kelton, Fraser and Miller, Luke and Simens, Maddie and Askell, Amanda and Welinder, Peter and Christiano, Paul and Leike, Jan and Lowe, Ryan},
  year         = {2022},
  eprint       = {2203.02155},
  archivePrefix= {arXiv},
  primaryClass = {cs.CL},
  doi          = {10.48550/arXiv.2203.02155}
}

@misc{lightman2023prm,
  title        = {Let's Verify Step by Step},
  author       = {Lightman, Hunter and Kosaraju, Vineet and Burda, Yura and Edwards, Harri and Baker, Bowen and Lee, Teddy and Leike, Jan and Schulman, John and Sutskever, Ilya and Cobbe, Karl},
  year         = {2023},
  eprint       = {2305.20050},
  archivePrefix= {arXiv},
  primaryClass = {cs.LG},
  doi          = {10.48550/arXiv.2305.20050}
}

@misc{cobbe2021gsm8k,
  title        = {Training Verifiers to Solve Math Word Problems},
  author       = {Cobbe, Karl and Kosaraju, Vineet and Bavarian, Mohammad and Chen, Mark and Jun, Heewoo and Kaiser, Lukasz and Plappert, Matthias and Tworek, Jerry and Hilton, Jacob and Nakano, Reiichiro and Hesse, Christopher and Schulman, John},
  year         = {2021},
  eprint       = {2110.14168},
  archivePrefix= {arXiv},
  primaryClass = {cs.LG},
  doi          = {10.48550/arXiv.2110.14168}
}

@inproceedings{wei2022cot,
  title     = {Chain-of-Thought Prompting Elicits Reasoning in Large Language Models},
  author    = {Wei, Jason and Wang, Xuezhi and Schuurmans, Dale and Bosma, Maarten and Ichter, Brian and Xia, Fei and Chi, Ed and Le, Quoc V. and Zhou, Denny},
  booktitle = {Advances in Neural Information Processing Systems (NeurIPS)},
  year      = {2022}
}

@inproceedings{lu2023mathvista,
  title        = {{MathVista}: Evaluating Mathematical Reasoning of Foundation Models in Visual Contexts},
  author       = {Lu, Pan and Bansal, Hritik and Xia, Tony and Liu, Jiacheng and Li, Chunyuan and Hajishirzi, Hannaneh and Cheng, Hao and Chang, Kai-Wei and Galley, Michel and Gao, Jianfeng},
  booktitle    = {The Twelfth International Conference on Learning Representations},
  year         = {2024},
  url          = {https://openreview.net/forum?id=KUNzEQMWU7}
}

@misc{bai2025qwen25vl,
  title        = {Qwen2.5-VL Technical Report},
  author       = {Bai, Shuai and Chen, Keqin and Liu, Xuejing and Wang, Jialin and Ge, Wenbin and Song, Sibo and Dang, Kai and Wang, Peng and Wang, Shijie and Tang, Jun and Zhong, Humen and Zhu, Yuanzhi and Yang, Mingkun and Li, Zhaohai and Wan, Jianqiang and Wang, Pengfei and Ding, Wei and Fu, Zheren and Xu, Yiheng and Ye, Jiabo and Zhang, Xi and Xie, Tianbao and Cheng, Zesen and Zhang, Hang and Yang, Zhibo and Xu, Haiyang and Lin, Junyang},
  year         = {2025},
  eprint       = {2502.13923},
  archivePrefix= {arXiv},
  primaryClass = {cs.CV},
  doi          = {10.48550/arXiv.2502.13923}
}

@misc{kimi_k15_2025,
  title        = {Kimi k1.5: Scaling Reinforcement Learning with LLMs},
  author       = {Kimi Team and Angang Du and Bofei Gao and Bowei Xing and Changjiu Jiang and Cheng Chen and Cheng Li and Chenjun Xiao and Chenzhuang Du and Chonghua Liao and Chuning Tang and Congcong Wang and Dehao Zhang and Enming Yuan and others},
  year         = {2025},
  eprint       = {2501.12599},
  archivePrefix= {arXiv},
  primaryClass = {cs.CL},
  note         = {arXiv preprint arXiv:2501.12599},
  url          = {https://arxiv.org/abs/2501.12599}
}

@inproceedings{lu2021intergps,
  title={Inter-GPS: Interpretable Geometry Problem Solving with Formal Language and Symbolic Reasoning},
  author={Lu, Pan and Gong, Ran and Jiang, Shibiao and Qiu, Liang and Huang, Siyuan and Liang, Xiaodan and Zhu, Song-Chun},
  booktitle={Proceedings of the 59th Annual Meeting of the Association for Computational Linguistics (ACL)},
  pages={6774--6786},
  year={2021}
}

@article{meng2025mmeureka,
  title={MM-Eureka: Exploring the Frontiers of Multimodal Reasoning with Rule-based Reinforcement Learning},
  author={Meng, Fanqing and Du, Lingxiao and Liu, Zongkai and Zhou, Zhixiang and Lu, Quanfeng and Fu, Daocheng and Han, Tiancheng and Shi, Botian and Wang, Wenhai and He, Junjun and Zhang, Kaipeng and Luo, Ping and Qiao, Yu and Zhang, Qiaosheng and Shao, Wenqi},
  journal={arXiv preprint arXiv:2503.07365},
  year={2025}
}

@inproceedings{zhang2024mathverse,
  title={MathVerse: Does Your Multi-modal LLM Truly See the Diagrams in Visual Math Problems?},
  author={Zhang, Renrui and Jiang, Dongzhi and Zhang, Yichi and Lin, Haokun and Guo, Ziyu and Qiu, Pengshuo and Zhou, Aojun and Lu, Pan and Chang, Kai-Wei and Gao, Peng and Li, Hongsheng},
  booktitle={European Conference on Computer Vision (ECCV)},
  pages={169--186},
  year={2024},
  organization={Springer}
}

@inproceedings{wang2024mathvision,
  title={Measuring Multimodal Mathematical Reasoning with the {MATH-Vision} Dataset},
  author={Wang, Ke and Pan, Junting and Shi, Weikang and Lu, Zimu and Ren, Houxing and Zhou, Aojun and Zhan, Mingjie and Li, Hongsheng},
  booktitle={Advances in Neural Information Processing Systems (NeurIPS)},
  volume={37},
  pages={95095--95169},
  year={2024}
}

@article{qiao2024wemath,
  title={{We-Math}: Does Your Large Multimodal Model Achieve Human-like Mathematical Reasoning?},
  author={Qiao, Runqi and Tan, Qiuna and Dong, Guanting and Wu, Minhui and Sun, Chong and Song, Xiaoshuai and GongQue, Zhuoma and Lei, Shanglin and Wei, Zhe and Zhang, Miaoxuan and Qiao, Runfeng and Zhang, Yifan and Zong, Xiao and Xu, Yida and Diao, Muxi and Bao, Zhimin and Li, Chen and Zhang, Honggang},
  journal={arXiv preprint arXiv:2407.01284},
  year={2024}
}

@misc{xiao2024logicvista,
  title={{LogicVista}: Multimodal {LLM} Logical Reasoning Benchmark in Visual Contexts},
  author={Xiao, Yijia and Sun, Edward and Liu, Tianyu and Wang, Wei},
  year={2024},
  eprint={2407.04973},
  archivePrefix={arXiv},
  primaryClass={cs.AI},
  doi={10.48550/arXiv.2407.04973}
}

@misc{yaowei2025easyr1,
  title={EasyR1: An Efficient, Scalable, Multi-Modality RL Training Framework},
  author={Zheng, Yaowei and Lu, Junting and Wang, Shenzhi and Feng, Zhangchi and Kuang, Dongdong and Xiong, Yuwen and Zhang, Richong},
  year={2025},
  publisher={GitHub},
  howpublished={\url{https://github.com/hiyouga/EasyR1}}
}

@article{leng2025mmr1,
  title={{MMR1}: Enhancing Multimodal Reasoning with Variance-Aware Sampling and Open Resources},
  author={Leng, Sicong and Wang, Jing and Li, Jiaxi and Zhang, Hao and Hu, Zhiqiang and Zhang, Boqiang and Jiang, Yuming and Zhang, Hang and Li, Xin and Bing, Lidong and Zhao, Deli and Lu, Wei and Rong, Yu and Sun, Aixin and Lu, Shijian},
  journal={arXiv preprint arXiv:2509.21268},
  year={2025}
}

@misc{guo2025seed,
  title={{Seed1.5-VL} Technical Report},
  author={Guo, Dong and others},
  year={2025},
  eprint={2505.07062},
  archivePrefix={arXiv},
  primaryClass={cs.CV},
  doi={10.48550/arXiv.2505.07062}
}

@misc{openai2025gpt41,
  title={Introducing {GPT-4.1} in the {API}},
  author={OpenAI},
  year={2025},
  howpublished={\url{https://openai.com/index/gpt-4-1/}}
}

@article{comanici2025gemini,
  title={Gemini 2.5: Pushing the Frontier with Advanced Reasoning, Multimodality, Long Context, and Next Generation Agentic Capabilities},
  author={Comanici, Gheorghe and Bieber, Eric and Schaekermann, Mike and Pasupat, Ice and Sachdeva, Noveen and Dhillon, Inderjit and Blistein, Marcel and Ram, Ori and others},
  journal={arXiv preprint arXiv:2507.06261},
  year={2025},
  url={https://arxiv.org/abs/2507.06261}
}

@misc{anthropic2025claude,
  title={Introducing {Claude 4}},
  author={Anthropic},
  year={2025},
  howpublished={\url{https://www.anthropic.com/research/claude-4}}
}

@article{xiao2025fast,
  title={{Fast-Slow Thinking GRPO} for Large Vision-Language Model Reasoning},
  author={Xiao, Wenyi and Gan, Leilei},
  journal={arXiv preprint arXiv:2504.18458},
  year={2025}
}

@article{chen2025vlaa,
  title={SFT or RL? An Early Investigation into Training R1-Like Reasoning Large Vision-Language Models},
  author={Chen, Hardy and Tu, Haoqin and Wang, Fali and Liu, Hui and Tang, Xianfeng and Du, Xinya and Zhou, Yuyin and Xie, Cihang},
  journal={arXiv preprint arXiv:2504.11468},
  year={2025},
  note={Introduces VLAA-Thinker}
}

@misc{deng2025openvlthinker,
      title={{OpenVLThinker}: Complex Vision-Language Reasoning via Iterative {SFT-RL} Cycles},
      author={Deng, Yihe and Bansal, Hritik and Yin, Fan and Peng, Nanyun and Wang, Wei and Chang, Kai-Wei},
      year={2025},
      eprint={2503.17352},
      archivePrefix={arXiv},
      primaryClass={cs.CV},
      url={https://arxiv.org/abs/2503.17352}, 
}

@article{wang2025thinklite,
  title={SoTA with Less: MCTS-Guided Sample Selection for Data-Efficient Visual Reasoning Self-Improvement},
  author={Wang, Xiyao and Yang, Zhengyuan and Feng, Chao and Lu, Hongjin and Li, Linjie and Lin, Chung-Ching and Lin, Kevin and Huang, Furong and Wang, Lijuan},
  journal={arXiv preprint arXiv:2504.07934},
  year={2025}
}

@inproceedings{wang2025vlrethinker,
  title={VL-Rethinker: Incentivizing Self-Reflection of Vision-Language Models with Reinforcement Learning},
  author={Wang, Haozhe and Qu, Chao and Huang, Zuming and Chu, Wei and Lin, Fangzhen and Chen, Wenhu},
  booktitle={Advances in Neural Information Processing Systems},
  year={2025},
  eprint={2504.08837},
  archivePrefix={arXiv},
  primaryClass={cs.CV},
  url={https://arxiv.org/abs/2504.08837}
}

@article{zha2025visiong1,
  title={Vision-G1: Towards General Vision Language Reasoning with Multi-Domain Data Curation},
  author={Zha, Yuheng and Zhou, Kun and Wu, Yujia and Wang, Yushu and Feng, Jie and Xu, Zhi and Hao, Shibo and Liu, Zhengzhong and Xing, Eric P and Hu, Zhiting},
  journal={arXiv preprint arXiv:2508.12680},
  year={2025}
}

@misc{zou2024dynamath,
      title={DynaMath: A Dynamic Visual Benchmark for Evaluating Mathematical Reasoning Robustness of Vision Language Models}, 
      author={Chengke Zou and Xingang Guo and Rui Yang and Junyu Zhang and Bin Hu and Huan Zhang},
      year={2024},
      eprint={2411.00836},
      archivePrefix={arXiv},
      primaryClass={cs.CV},
      doi={10.48550/arXiv.2411.00836},
      url={https://arxiv.org/abs/2411.00836}, 
}

@inproceedings{yue2024mmmu,
  title={{MMMU}: A Massive Multi-discipline Multimodal Understanding and Reasoning Benchmark for Expert {AGI}},
  author={Yue, Xiang and Ni, Yuansheng and Zhang, Kai and Zheng, Tianyu and Liu, Ruoqi and Zhang, Ge and Stevens, Samuel and Jiang, Dongfu and Ren, Weiming and Sun, Yuxuan and Wei, Cong and Yu, Botao and Yuan, Ruibin and Sun, Renliang and Yin, Ming and Zheng, Boyuan and Yang, Zhenzhu and Liu, Yibo and Huang, Wenhao and Sun, Huan and Su, Yu and Chen, Wenhu},
  booktitle={Proceedings of the IEEE/CVF Conference on Computer Vision and Pattern Recognition},
  pages={9556--9567},
  year={2024}
}

@article{yue2024mmmupro,
  title={{MMMU-Pro}: A More Robust Multi-discipline Multimodal Understanding Benchmark},
  author={Yue, Xiang and Zheng, Tianyu and Ni, Yuansheng and Wang, Yubo and Zhang, Kai and Tong, Shengbang and Sun, Yuxuan and Yu, Botao and Zhang, Ge and Sun, Huan and Su, Yu and Chen, Wenhu and Neubig, Graham},
  journal={arXiv preprint arXiv:2409.02813},
  year={2024}
}

@article{wang2024charxiv,
  title={{CharXiv}: Charting Gaps in Realistic Chart Understanding in Multimodal {LLMs}},
  author={Wang, Zirui and Xia, Mengzhou and He, Luxi and Chen, Howard and Liu, Yitao and Zhu, Richard and Liang, Kaiqu and Wu, Xindi and Liu, Haotian and Malladi, Sadhika and Chevalier, Alexis and Arora, Sanjeev and Chen, Danqi},
  journal={Advances in Neural Information Processing Systems},
  volume={37},
  pages={113569--113697},
  year={2024}
}

@article{chen2024mmstar,
title={Are We on the Right Way for Evaluating Large Vision-Language Models?},
author={Chen, Lin and Li, Jinsong and Dong, Xiaoyi and Zhang, Pan and Zang, Yuhang and Chen, Zehui and Duan, Haodong and Wang, Jiaqi and Qiao, Yu and Lin, Dahua and Zhao, Feng},
journal={arXiv preprint arXiv:2403.20330},
year={2024}
}

@article{shen2025satori,
  title={SATORI-R1: Incentivizing Multimodal Reasoning through Explicit Visual Anchoring},
  author={Shen, Chuming and Wei, Wei and Qu, Xiaoye and Cheng, Yu},
  journal={arXiv preprint arXiv:2505.19094},
  year={2025}
}

@article{zhu2025retrv,
  title={Retrv-R1: A Reasoning-Driven MLLM Framework for Universal and Efficient Multimodal Retrieval},
  author={Zhu, Lanyun and Ji, Deyi and Chen, Tianrun and Wu, Haiyang and Wang, Shiqi},
  journal={arXiv preprint arXiv:2510.02745},
  year={2025}
}

@article{wu2025quantile,
  title={Quantile Advantage Estimation: Stabilizing {RLVR} for {LLM} Reasoning},
  author={Wu, Junkang and Huang, Kexin and Wu, Jiancan and Zhang, An and Wang, Xiang and He, Xiangnan},
  journal={arXiv preprint arXiv:2509.22611},
  year={2025}
}

@article{li2025rorecomp,
  title={RoRecomp: Enhancing Reasoning Efficiency via Rollout Response Recomposition in Reinforcement Learning},
  author={Li, Gang and Qin, Yulei and Tan, Xiaoyu and Yang, Dingkang and Shi, Yuchen and Xu, Zihan and Li, Xiang and Sun, Xing and Li, Ke},
  journal={arXiv preprint arXiv:2509.25958},
  year={2025}
}

@article{zeng2025shrinking,
  title={Shrinking the Variance: Shrinkage Baselines for Reinforcement Learning with Verifiable Rewards},
  author={Zeng, Guanning and Zhou, Zhaoyi and Arora, Daman and Zanette, Andrea},
  journal={arXiv preprint arXiv:2511.03710},
  year={2025}
}

@article{le2025token,
  title={Sharpness-Guided Group Relative Policy Optimization via Probability Shaping},
  author={Le, Tue and Van, Linh Ngo and Le, Trung},
  journal={arXiv preprint arXiv:2511.00066},
  year={2025}
}

@article{yang2025entropic,
  title={EntroPIC: Towards Stable Long-Term Training of LLMs via Entropy Stabilization with Proportional-Integral Control},
  author={Yang, Kai and Xu, Xin and Chen, Yangkun and Liu, Weijie and Lyu, Jiafei and Lin, Zichuan and Ye, Deheng and Yang, Saiyong},
  journal={arXiv preprint arXiv:2511.15248},
  year={2025}
}

@article{bai2025mgrpo,
  title={M-GRPO: Stabilizing Self-Supervised Reinforcement Learning for Large Language Models with Momentum-Anchored Policy Optimization},
  author={Bai, Bizhe and Wu, Hongming and Ye, Peng and Chen, Tao},
  journal={arXiv preprint arXiv:2512.13070},
  year={2025}
}

@inproceedings{malinin2018predictive,
  title={Predictive Uncertainty Estimation via Prior Networks},
  author={Malinin, Andrey and Gales, Mark},
  booktitle={Advances in Neural Information Processing Systems (NeurIPS)},
  volume={31},
  pages={7047--7058},
  year={2018}
}

@inproceedings{feng2026video,
  title={Video-R1: Reinforcing Video Reasoning in MLLMs},
  author={Feng, Kaituo and Gong, Kaixiong and Li, Bohao and Guo, Zonghao and Wang, Yibing and Peng, Tianshuo and Wu, Junfei and Zhang, Xiaoying and Wang, Benyou and Yue, Xiangyu},
  booktitle={Advances in Neural Information Processing Systems},
  volume={38},
  pages={99114--99137},
  year={2025}
}

@article{chen2025advancing,
  title={Advancing multimodal reasoning: From optimized cold start to staged reinforcement learning},
  author={Chen, Shuang and Guo, Yue and Su, Zhaochen and Li, Yafu and Wu, Yulun and Chen, Jiacheng and Chen, Jiayu and Wang, Weijie and Qu, Xiaoye and Cheng, Yu},
  journal={arXiv preprint arXiv:2506.04207},
  year={2025}
}

@inproceedings{chen2026ares,
  title={Ares: Multimodal adaptive reasoning via difficulty-aware token-level entropy shaping},
  author={Chen, Shuang and Guo, Hangyu and Ye, Yimeng and Huang, Shijue and Hu, Wenbo and Chen, Jiayu and Zhang, Manyuan and Li, Haoxi and Guo, Song and Peng, Nanyun Violet},
  booktitle={International Conference on Learning Representations},
  volume={2026},
  pages={147474--147512},
  year={2026}
}

@article{zhang2025critique,
  title={Critique-GRPO: Advancing LLM Reasoning with Natural Language and Numerical Feedback},
  author={Zhang, Xiaoying and Sun, Hao and Zhang, Yipeng and Feng, Kaituo and Yang, Chao and Meng, Helen},
  journal={arXiv preprint arXiv:2506.03106},
  year={2025}
}

@inproceedings{wu2026reinforcing,
  title={Reinforcing Spatial Reasoning in Vision-Language Models with Interwoven Thinking and Visual Drawing},
  author={Wu, Junfei and Guan, Jian and Feng, Kaituo and Liu, Qiang and Wu, Shu and Wang, Liang and Wu, Wei and Tan, Tieniu},
  booktitle={Advances in Neural Information Processing Systems},
  volume={38},
  pages={143297--143330},
  year={2025}
}

@inproceedings{feng2026onethinker,
  title={Onethinker: All-in-one reasoning model for image and video},
  author={Feng, Kaituo and Zhang, Manyuan and Li, Hongyu and Fan, Kaixuan and Chen, Shuang and Jiang, Yilei and Zheng, Dian and Sun, Peiwen and Zhang, Yiyuan and Sun, Haoze and others},
  booktitle={Proceedings of the IEEE/CVF Conference on Computer Vision and Pattern Recognition},
  pages={5432--5443},
  year={2026}
}

@article{chen2026unify,
  title={Unify-agent: A unified multimodal agent for world-grounded image synthesis},
  author={Chen, Shuang and Shou, Quanxin and Chen, Hangting and Zhou, Yucheng and Feng, Kaituo and Hu, Wenbo and Zhang, Yi-Fan and Lin, Yunlong and Huang, Wenxuan and Song, Mingyang and others},
  journal={arXiv preprint arXiv:2603.29620},
  year={2026}
}

@article{feng2026gen,
  title={Gen-searcher: Reinforcing agentic search for image generation},
  author={Feng, Kaituo and Zhang, Manyuan and Chen, Shuang and Lin, Yunlong and Fan, Kaixuan and Jiang, Yilei and Li, Hongyu and Zheng, Dian and Wang, Chenyang and Yue, Xiangyu},
  journal={arXiv preprint arXiv:2603.28767},
  year={2026}
}

@article{chen2026opensearch,
  title={Opensearch-vl: An open recipe for frontier multimodal search agents},
  author={Chen, Shuang and Feng, Kaituo and Chen, Hangting and Huang, Wenxuan and Dai, Dasen and Shou, Quanxin and Lin, Yunlong and Yue, Xiangyu and Gao, Shenghua and Pang, Tianyu},
  journal={arXiv preprint arXiv:2605.05185},
  year={2026}
}

@article{huang2026vision,
  title={Vision-DeepResearch: Incentivizing DeepResearch Capability in Multimodal Large Language Models},
  author={Huang, Wenxuan and Zeng, Yu and Wang, Qiuchen and Fang, Zhen and Cao, Shaosheng and Chu, Zheng and Yin, Qingyu and Chen, Shuang and Yin, Zhenfei and Chen, Lin and others},
  journal={arXiv preprint arXiv:2601.22060},
  year={2026}
}

@article{zeng2026vision,
  title={Vision-DeepResearch Benchmark: Rethinking Visual and Textual Search for Multimodal Large Language Models},
  author={Zeng, Yu and Huang, Wenxuan and Fang, Zhen and Chen, Shuang and Shen, Yufan and Cai, Yishuo and Wang, Xiaoman and Yin, Zhenfei and Chen, Lin and Chen, Zehui and others},
  journal={arXiv preprint arXiv:2602.02185},
  year={2026}
}

@misc{fang2026videodeepresearchnextgenerationmultimodaldeepresearch,
      title={Video-DeepResearch: Towards the Next-Generation Multimodal Deepresearch Agent}, 
      author={Zhen Fang and Yu Zeng and Wenxuan Huang and Yiming Zhao and Shiting Huang and Tianfei Ren and Qi Lu and Qingnan Ren and Qisheng Su and Lionel Z. Wang and Qingyu Yin and Shuang Chen and Zehui Chen and Lin Chen and Zhenfei Yin and Yao Hu and Shaohui Lin and Wanli Ouyang and Shaosheng Cao and Feng Zhao},
      year={2026},
      eprint={2608.03979},
      archivePrefix={arXiv},
      primaryClass={cs.CV},
      url={https://arxiv.org/abs/2608.03979}, 
}

\appendix

\section{Appendix}
\label{sec:appendix}

\subsection{Hyperparameter Sensitivity}
\label{sec:appendix_sensitivity}
We conducted a sensitivity analysis on the key hyperparameters of HSR: the stream split ratio (Structure/Natural) and the percentile thresholds ($\tau$).
\begin{itemize}
    \item \textbf{Stream Ratio:} We swept the Structured Stream ratio from 0\% (Pure GRPO) to 100\%. We observed that a ratio of 50\% provides the optimal balance. Higher ratios (>75\%) led to overfitting on the replay buffer, while lower ratios (<25\%) failed to prevent entropy collapse.
    \item \textbf{Path-Entropy Thresholds:} We tested $\tau_{low} \in \{5, 10, 20\}$th percentile. Results indicate that stringent thresholds (5th percentile) overly restrict the anchor pool, leading to buffer depletion, while loose thresholds (20th percentile) allow noisy samples into the anchor set, degrading stability. The 10th/90th percentile setting proved robust across all examined benchmarks.
\end{itemize}

\subsection{Algorithmic Implementation Details}
\label{app:algo_details}

In this section, we provide a step-by-step commentary on the training procedure outlined in Algorithm \ref{alg:stable_mm_r1}. The framework extends standard Group Relative Policy Optimization (GRPO) by introducing a dynamic curriculum (Phase 0) and a structured replay mechanism (Phase 1 \& 2).

\paragraph{Initialization and Inputs.}
The algorithm requires a curated multimodal instruction dataset $\mathcal{D}$, the policy model $\pi_\theta$ to be optimized, and an offline hint cache $\mathcal{C}_{hint}$ generated in advance by a frozen proprietary teacher model $\pi_{seed}$ (Seed-1.5-VL). For each query, we use a per-query rollout group size of $G=N=16$. This group is subsequently partitioned between the structured and natural streams.

\paragraph{Phase 0: Dynamic Query Handling (Lines 3--12).}
Standard RL fine-tuning often wastes computational resources on queries that are currently intractable for the model.
\begin{itemize}
    \item \textbf{Pilot Assessment:} For each sampled query $x$, we generate $N$ pilot rollouts to estimate the empirical success rate $\hat{p}(x)$.
    \item \textbf{Intervention:} If $\hat{p}(x)=0$ (Line 6), the query is classified as ``Impossible.'' Rather than discarding these samples, we retrieve its precomputed visual or cognitive hint $h$ from the offline cache $\mathcal{C}_{hint}$. The prompt is augmented ($x' \leftarrow x \oplus h$) during Phase 0 curriculum construction, and rollouts are then \textit{regenerated} (Line 10). This mechanism converts selected ``Impossible'' queries into ``Learnable'' ones (the Distillation Zone) prior to gradient computation.
\end{itemize}

\paragraph{Phase 1: Path-Entropy-Based Stratification (Lines 13--24).}
Upon finalizing the rollouts for the current optimization step, we compute the reward $r$ and Path-Entropy score $H_{\mathrm{path}}(y|x)$ for every sample.
\begin{itemize}
    \item \textbf{Dynamic Thresholding:} We pool the Path-Entropy scores of all rollouts generated for the current global batch and calculate their $10^{th}$ and $90^{th}$ percentiles ($\tau_{low}, \tau_{high}$). The thresholds are therefore batch-adaptive and use no samples from previous optimization steps.
    \item \textbf{Buffer Allocation:} Samples are stratified into three mutually exclusive buffers: $\mathcal{B}_{anc}$ (high confidence, correct), $\mathcal{B}_{neg}$ (high confidence, incorrect), and $\mathcal{B}_{epi}$ (low confidence, correct). Note that ``Confusion'' samples ($r=0, H_{\mathrm{path}} > \tau_{high}$) are implicitly filtered out, as they do not meet the criteria for any buffer.
\end{itemize}

\paragraph{Phase 2: Hybrid Group Assembly (Lines 25--32).}
To construct the final per-query optimization group $G_{final}$, we partition the $G=16$ rollouts into two streams that balance stability and exploration:
\begin{itemize}
    \item \textbf{Stream A (Structured):} We explicitly sample $50\%$ of the total budget $G$ from the sorted buffers. We enforce a composition of $25\%$ Epiphanies (learning targets), $12.5\%$ Anchors (stability), and $12.5\%$ Hard Negatives (contrast) whenever the corresponding samples are available. This structured injection yields non-zero empirical reward variance when both outcome classes are present; otherwise, the unavailable quota falls back to natural on-policy sampling.
    \item \textbf{Stream B (Natural):} The remaining $50\%$ is sampled uniformly from the valid sample pool (excluding noise). This stream acts as a regularizer, ensuring the policy does not overfit to the artificial topology of Stream A.
\end{itemize}

\paragraph{Optimization and Lifecycle (Lines 33--36).}
Finally, gradients are computed using the combined group $G_{final}$. A critical feature of our implementation is the Intra-Step Replay Scope (Line 36). Unlike off-policy methods with persistent replay buffers, our buffers $\mathcal{B}_{anc, neg, epi}$ are cleared immediately after every gradient step. This ensures that all optimization samples are drawn from the current policy $\pi_\theta$, maintaining the on-policy consistency of the GRPO objective without requiring corrections for stale data.

\begin{algorithm}[H]
  \caption{Stable-MM-R1 Unified Training Framework}
  \label{alg:stable_mm_r1}
  \small
  \begin{algorithmic}[1]
    \Require Dataset $\mathcal{D}$; Current Policy $\pi_\theta$; Offline Hint Cache $\mathcal{C}_{hint}$; 
            Pilot Rollout Count and Final Group Size $N=G=16$;
            Quantile thresholds $\delta_{low}=0.1, \delta_{high}=0.9$
    \Ensure  Optimized Policy parameters $\theta$
    \While{not converged}
        \State Sample batch query $x \sim \mathcal{D}$
        \Statex \Comment{\textbf{Phase 0: Potential-Aware Query Mining (PAQM)}}
        \State $\mathcal{Y}_{pilot} \gets \{y_1, \dots, y_N\} \sim \pi_\theta(\cdot|x)$ \Comment{Generate pilot rollouts}
        \State $\hat{p}(x) \gets \frac{1}{N}\sum_{i=1}^N r(y_i)$ \Comment{Estimate empirical success rate}
        
        \If{$\hat{p}(x) = 0$} 
            \Statex \Comment{\textit{   a. Apply Modality-Decoupled Diagnosis for Impossible Queries}}
            \State $h \gets \textsc{LookupHint}(x, \mathcal{C}_{hint})$ \Comment{Retrieve precomputed offline hint}
            \State $x' \gets x \oplus h$ \Comment{Inject hint during curriculum construction}
            \State $\mathcal{Y} \gets \textsc{Resample}(\pi_\theta, x', N)$ \Comment{Regenerate rollouts with hint}
        \Else
            \State $\mathcal{Y} \gets \mathcal{Y}_{pilot}$
        \EndIf
        \Statex \Comment{\textbf{Phase 1: Diagnostic Path-Entropy--Reward Stratification}}
        \State $\mathcal{H}_t \gets \{H_{\mathrm{path}}(y|x) \mid y \in \mathcal{Y}_{\mathrm{batch},t}\}$ \Comment{Pool current global-batch rollouts}
        \State $\tau_{low} \gets \textsc{Percentile}(\mathcal{H}_t, \delta_{low}); \quad \tau_{high} \gets \textsc{Percentile}(\mathcal{H}_t, \delta_{high})$
        
        \State $\mathcal{B}_{anc} \gets \emptyset; \quad \mathcal{B}_{neg} \gets \emptyset; \quad \mathcal{B}_{epi} \gets \emptyset$
        \ForAll{$y \in \mathcal{Y}$}
            \If{$r(y)=1 \land H_{\mathrm{path}}(y|x) < \tau_{low}$} 
                \State $\mathcal{B}_{anc} \gets \mathcal{B}_{anc} \cup \{y\}$ \Comment{Stability Anchor}
            \ElsIf{$r(y)=0 \land H_{\mathrm{path}}(y|x) < \tau_{low}$}
                \State $\mathcal{B}_{neg} \gets \mathcal{B}_{neg} \cup \{y\}$ \Comment{Hard Negative}
            \ElsIf{$r(y)=1 \land H_{\mathrm{path}}(y|x) > \tau_{high}$}
                \State $\mathcal{B}_{epi} \gets \mathcal{B}_{epi} \cup \{y\}$ \Comment{Epiphany (Target)}
            \EndIf
        \EndFor
        \Statex \Comment{\textbf{Phase 2: Hybrid Group Assembly}}
        \Statex \Comment{\textit{   b. Stream A: Structured Stabilization ($50\%$ Budget)}}
        \State $S_{epi} \gets \textsc{Sample}(\mathcal{B}_{epi}, 0.25G)$
        \State $S_{anc} \gets \textsc{Sample}(\mathcal{B}_{anc}, 0.125G)$
        \State $S_{neg} \gets \textsc{Sample}(\mathcal{B}_{neg}, 0.125G)$
        \State $G_{struct} \gets S_{epi} \cup S_{anc} \cup S_{neg}$
        
        \Statex \Comment{\textit{   c. Stream B: Natural Generalization ($50\%$ Budget)}}
        \State $G_{nat} \gets \textsc{Sample}(\mathcal{Y} \setminus \text{Noise}, 0.5G)$
        
        \Statex \Comment{\textbf{Optimization Update (Intra-Step Scope)}}
        \State $G_{final} \gets G_{struct} \cup G_{nat}$
        \State $\theta \gets \textsc{GRPO\_Update}(\theta, G_{final})$
        \State $\text{Clear}(\mathcal{B}_{anc}, \mathcal{B}_{neg}, \mathcal{B}_{epi})$ \Comment{Buffers are cleared immediately after step}
    \EndWhile
  \end{algorithmic}
\end{algorithm}

\subsection{Conditional Variance Analysis of Structured Groups}
\label{sec:appendix_theory_detailed}

This section analyzes the conditional effect of structured group composition in sparse-reward Group Relative Policy Optimization (GRPO). The analysis motivates HSR but does not assert a non-zero update for every optimization step.

\subsubsection{Degenerate Groups in Standard GRPO}

Let $\mathcal{G} = \{y_1, \dots, y_G\}$ be a group of outputs sampled from policy $\pi_\theta(\cdot|x)$. The GRPO gradient estimator is defined as:
\begin{equation}
    \hat{g}_{GRPO} = \frac{1}{G} \sum_{i=1}^G \frac{r(y_i) - \bar{r}}{\hat{\sigma} + \epsilon} \nabla_\theta \log \pi_\theta(y_i|x),
\end{equation}
where $\bar{r}$ and $\hat{\sigma}$ are the empirical mean and standard deviation of rewards within the group.

\paragraph{Degenerate-group probability.}
In a binary reward setting $r \in \{0, 1\}$, let $p_\theta(x)$ be the true success rate of the policy. The probability of sampling a \textit{degenerate group}, in which all rewards are identical, is
\begin{equation}
    P_{degen} = p_\theta(x)^G + (1-p_\theta(x))^G.
\end{equation}
In such groups, $\hat{\sigma}=0$, so the group-relative advantages vanish or become numerically ill-conditioned depending on the implementation. When $p_\theta(x)$ is close to 0 or 1, $P_{degen}$ increases, making zero-signal groups more likely under random sampling.

\subsubsection{Conditional Variance of Structured Groups}

Conditional on the relevant buffers being non-empty, HSR's structured stream includes at least $k_{anc}$ positive samples (Stability Anchors) and $k_{neg}$ negative samples (Hard Negatives). Let $G_{struct}$ be the structured-group size.

\paragraph{Conditional lower bound.}
If $k_{anc} \ge 1$ and $k_{neg} \ge 1$, the empirical reward variance of the structured group is strictly positive. Let $N_+ \ge k_{anc}$ and $N_- \ge k_{neg}$ denote the numbers of positive and negative samples, with $N_+ + N_- = G_{struct}$. Then
\begin{equation}
\begin{aligned}
    \hat{\sigma}^2
    &= \frac{1}{G_{struct}} \left[N_+(1-\bar{r})^2 + N_-(0-\bar{r})^2\right] \\
    &= \frac{N_+N_-}{G_{struct}^2} > 0.
\end{aligned}
\end{equation}
Thus, the structured group has non-zero reward variance whenever both outcome classes are available. If either quota cannot be satisfied, HSR releases the unavailable structured allocation and fills it through natural on-policy sampling. The bound is therefore conditional rather than a guarantee for every optimization step.

\subsubsection{Distribution-Shift Trade-off}

Structured sampling changes the empirical training distribution by emphasizing anchors, hard negatives, and high-Path-Entropy successes. We treat this as a deliberate distribution shift toward informative boundary cases rather than claiming that the resulting estimator is unbiased. Stream B retains natural on-policy samples within every optimization step, while the empirical training-dynamics results evaluate whether the resulting trade-off improves stability in practice.

\end{document}